\documentclass[letterpaper]{article} 
\usepackage{aaai2026}  
\usepackage{times}  
\usepackage{helvet}  
\usepackage{courier}  
\usepackage[hyphens]{url}  
\usepackage{graphicx} 
\usepackage{natbib}  
\usepackage{caption} 
\usepackage{algorithm}
\usepackage{algorithmic}

\usepackage{amsmath, amssymb}
\usepackage{multirow}
\usepackage{cleveref}
\usepackage{xcolor}

\usepackage{booktabs}
\usepackage{array}
\usepackage{caption}
\usepackage{multirow}
\usepackage{adjustbox} 


\usepackage{newfloat}
\usepackage{listings}
\DeclareCaptionStyle{ruled}{labelfont=normalfont,labelsep=colon,strut=off} 
\floatstyle{ruled}
\newfloat{listing}{tb}{lst}{}
\floatname{listing}{Listing}
\title{Trusted Multi-view Learning for Long-tailed Classification}
\author {
    Chuanqing Tang\textsuperscript{\rm 1,\rm 2},
    Yifei Shi\textsuperscript{\rm 1},
    Guanghao Lin\textsuperscript{\rm 1},
    Lei Xing\textsuperscript{\rm 3},
    Long Shi\textsuperscript{\rm 1,\rm 2}\thanks{Corresponding author.}
}
\affiliations {
    \textsuperscript{\rm 1}School of Computing and Artificial Intelligence, Southwestern University of Finance and Economics, Chengdu, China\\
    \textsuperscript{\rm 2}Artificial Intelligence and Digital Finance Key Laboratory of Sichuan Province, Chengdu, China\\
    \textsuperscript{\rm 3}Xi’an Jiaotong University, Xi’an, China\\
    223081200012@smail.swufe.edu.cn, 42211176@smail.swufe.edu.cn, 42211096@smail.swufe.edu.cn,\\
    xingl@xjtu.edu.cn,
    shilong@swufe.edu.cn
}

\usepackage{bibentry}

\begin{document}

\maketitle

\begin{abstract}
Class imbalance has been extensively studied in single-view scenarios; however, addressing this challenge in multi-view contexts remains an open problem, with even scarcer research focusing on trustworthy solutions. In this paper, we tackle a particularly challenging class imbalance problem in multi-view scenarios: long-tailed classification. We propose TMLC, a Trusted Multi-view Long-tailed Classification framework, which makes contributions on two critical aspects: opinion aggregation and pseudo-data generation. Specifically, inspired by Social Identity Theory, we design a group consensus opinion aggregation mechanism that guides decision-making toward the direction favored by the majority of the group. In terms of pseudo-data generation, we introduce a novel distance metric to adapt SMOTE for multi-view scenarios and develop an uncertainty-guided data generation module that produces high-quality pseudo-data, effectively mitigating the adverse effects of class imbalance. Extensive experiments on long-tailed multi-view datasets demonstrate that our model is capable of achieving superior performance. The code is released at https://github.com/cncq-tang/TMLC.
\end{abstract}


\section{Introduction}

Class imbalance refers to a situation in which the distribution of classes in a dataset is non-uniform, characterized by significant or even extreme disparities~\cite{kang2020exploring}. This phenomenon, widely observed in real-world scenarios such as fraud detection~\cite{zhang2021fraudre} and medical image analysis~\cite{ju2021relational}, often arises due to limitations or biases in the data collection process. In the machine learning community, class imbalance is recognized as a critical research challenge, as models trained on imbalanced data tend to favor the majority class while neglecting the minority class~\cite{cao2019learning,tan2020equalization}. This can result in suboptimal performance, particularly for those predictions where accurate classification of the minority class is essential. 

Research on class imbalance mainly follows three technical routes: re-sampling~\cite{liu2008exploratory}, cost-sensitive learning~\cite{hsieh2021droploss}, and data augmentation~\cite{liu2020deep}. Re-sampling methods aim to balance class distributions by either over-sampling minority classes or under-sampling majority classes, with two well-known techniques being SMOTE~\cite{chawla2002smote} and Tomek Links~\cite{4309452}. In contrast to re-sampling, which focuses on data generation, cost-sensitive learning addresses class imbalance by assigning higher misclassification costs to minority classes through weighted loss functions~\cite{cui2019class,ren2020balanced} or specialized algorithms~\cite{cao2019learning,khan2019striking}. This approach offers flexibility but often requires careful tuning of cost parameters. Data augmentation, on the other hand, highlights the diversity of minority class samples using techniques such as generative models~\cite{ji2024tltscore} or feature-space transformations~\cite{wang2017learning}. It is worth noting that, although both re-sampling and data augmentation methods essentially fall under pseudo-data generation approaches, they differ significantly in their mechanisms: re-sampling focuses on adjusting data distribution, whereas data augmentation emphasizes enriching data diversity.

The aforementioned studies on class imbalance have been conducted in single-view scenarios, where decision-making relies on a single modality of data. However, real-world applications often involve multi-view scenarios~\cite{shi2024enhanced, shi2024tensor}, in which multiple data sources are available for analysis. For instance, in medical diagnosis~\cite{ren2024deep}, multi-view data may include X-ray images, MRI scans, and other modalities, each offering unique and complementary insights into patient conditions. To the best of our knowledge, addressing class imbalance in such multi-view settings remains an open and under-explored challenge, which demands innovative solutions and deeper investigation.

\begin{figure*}[t]
  \centering
   \includegraphics[width=0.8\linewidth]{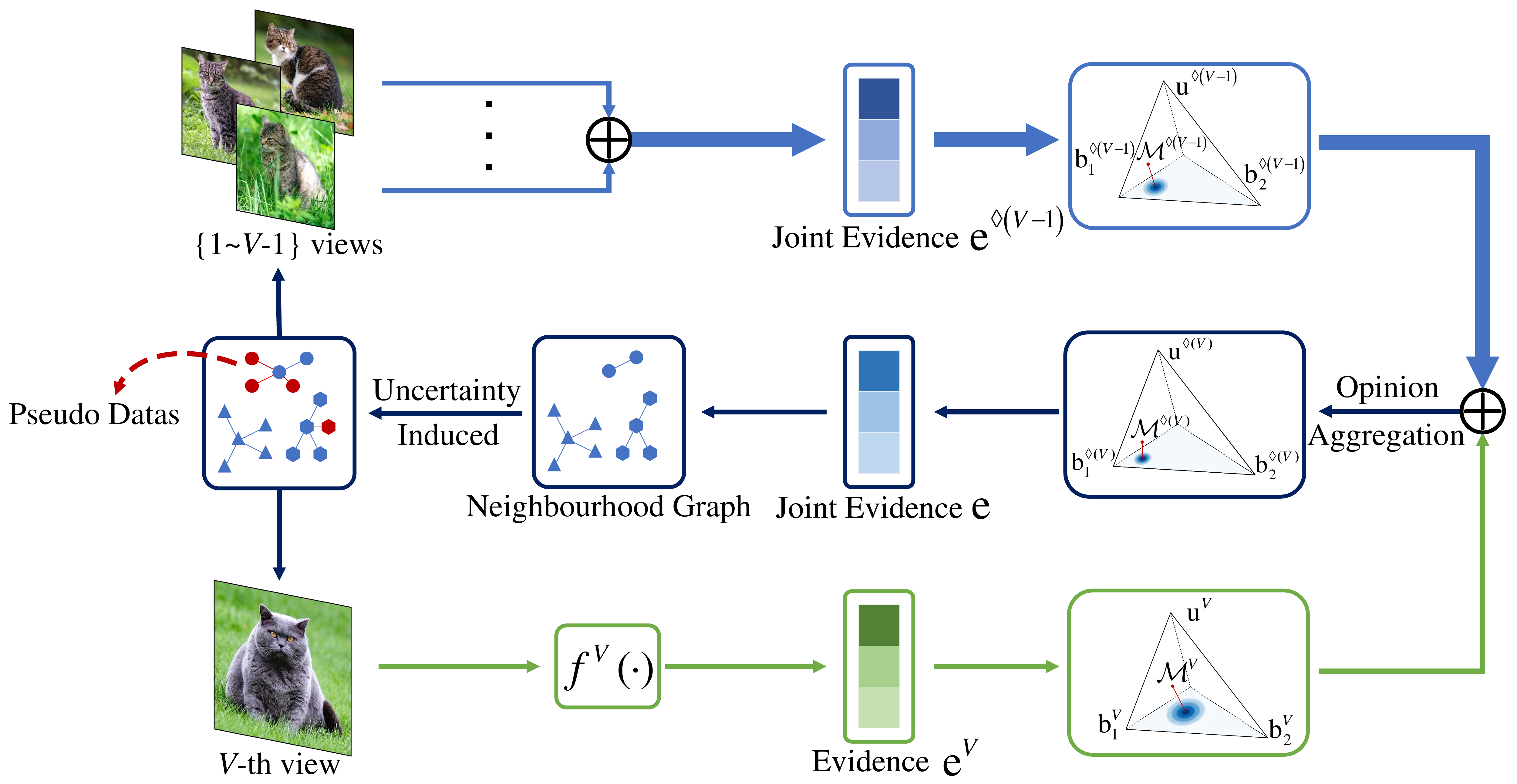}
   \caption{Illustration of TMLC. Assuming that the opinions from the first $V-1$ views have been aggregated, we now focus on combining the resulting joint opinion $\boldsymbol{e}^{\Diamond(V-1)}$ with the opinion $\boldsymbol{e}^{V}$ from the $V$-th view. In this process, a larger weight is assigned to $\boldsymbol{e}^{\Diamond(V-1)}$ which respresents the group joint opinion (bold arrow line), while a smaller weight is assigned to the individual opinion $\boldsymbol{e}^{V}$ (thin arrow line). Subsequently, we propose an uncertainty-based sample generation scheme to ensure the production of reliable pseudo-data for the minority class.}
   \label{fig:TMLC}
\end{figure*}

In this paper, we address a particularly challenging class imbalance problem—long-tailed distribution class imbalance—in multi-view scenarios. \textit{We aim to answer two critical questions: (1) How to tackle the challenge of multi-view long-tailed classification, and (2) how to ensure reliable decision-making?} To this end, we propose TMLC, a Trusted Multi-View Long-Tailed Classification Framework , as shown in Fig. \ref{fig:TMLC}. TMLC is designed to achieve trustworthy classification for multi-view long-tailed data by leveraging a deep evidence learning-inspired paradigm~\cite{sensoy2018evidential} and adopting a oversampling-based scheme. Specifically, we make significant advancements in opinion aggregation and pseudo-data generation, which respectively enhance the reliability of the trustworthy framework and improve the effectiveness of the oversampling process. The main contributions of this work are as follows:

\begin{itemize}
    \item Motivated from Social Identity Theory~\cite{hogg2016social}, we design a group consensus opinion aggregation mechanism that guides decision-making toward the direction favored by the majority of the group, thereby preventing negative individual opinions from dominating the final outcome.
    \item To address the limitation of SMOTE, which cannot be directly applied to multi-view settings, we introduce a novel distance metric based on joint subjective evidence and propose an uncertainty-based sample generation scheme that enables the production of high-quality pseudo-data.
    \item Extensive experiments on long-tailed multi-view datasets demonstrated that TMLC outperforms serveral state-of-art baselines.
\end{itemize}

\section{Related Work}



\textbf{Long-tailed Learning.} Traditional long-tailed learning methods include re-sampling~\cite{han2005borderline,liu2008exploratory}, cost-sensitive learning~\cite{cui2019class,cao2019learning,wang2021adaptive} and data augmentation~\cite{chu2020feature,kim2020m2m,yin2019feature}. As one of the most widely used techniques, Re-sampling rebalances classes by adjusting the number of samples per class. SMOTE~\cite{chawla2002smote} enhances the representation of minority classes by generating synthetic samples. Decoupling~\cite{kang2019decoupling} identifies effective sampling strategies for standard model training. DCL~\cite{wang2019dynamic} and LOCE~\cite{feng2021exploring} propose adaptive sampling strategies. Balanced Meta-Softmax~\cite{ren2020balanced} employs meta-learning to estimate optimal sampling rates for different classes. SimCal~\cite{wang2020devil} uses a novel two-level class-balanced sampling strategy that combines image-level and instance-level. However, these methods do not fully explore the uncertainty of samples during the sampling phase.

\noindent \textbf{Uncertainty-induced Trusted Learning.} Inspired by evidential deep learning~\cite{sensoy2018evidential}, which serves as a prominent paradigm for uncertainty estimation, Trusted Multi-view Classification (TMC) has recently gained particular attention. TMC leverages the Dirichlet distribution to model class probabilities and integrates Dempster-Shafer theory to achieve reliable classification results~\cite{han2022trusted}. Based on TMC, numerous extensions have been proposed to tackle specific challenges, involving opinion aggregation ~\cite{liu2022trusted, xu2024reliable, zheng2023multi}, noisy inputs and labels~\cite{xu2024trusted, shi2024generalized}, semantic ambiguity~\cite{liu2024dynamic}, etc. Recent advancements in trusted learning have expanded its research to areas such as weakly supervised learning~\cite{wang2024trusted, hu2025self}, robust learning ~\cite{zhou2023rtmc, zhou2023calm, du2023bridging, deng2025trustworthy}, and multimodal applications ~\cite{huang2025multi, li2025deep}.

\section{Methodology}

In this section, we first provide the problem definition and theoretical foundation on evidential deep learning. Subsequently, we detail our key contributions, including the group consensus opinion aggregation mechanism and the uncertainty-based sample generation scheme.


\subsection{Problem Definition and Preliminaries}

\subsubsection{Problem Definition}

A long-tailed multi-view dataset typically contains an imbalanced training set and a balanced test set. For a $K$ classification problem, we define the multi-view data as $\{\boldsymbol{X}^{v}\}^V_{v=1}$ and the corresponding ground-truth labels as $\{\boldsymbol{\mathrm{y}}_{n} \}_{n=1}^{N}$ with $V$ views and $N$ samples, where $\boldsymbol{X}^{v} =[\boldsymbol{x}^v_1,\boldsymbol{x}^v_2,\cdots,\boldsymbol{x}^v_N] \in \mathbb{R}^{d_v\times N}$ represents the $v$-th view data with $d_v$ dimension. For each view data $\boldsymbol{X}^{v}$, the distribution of classes is highly imbalanced, with tail classes having significantly fewer samples compared to head classes. The goal of TMLC is to address the challenges caused by imbalanced class distributions.    

\subsubsection{Evidential Deep Learning}

Evidential Deep Learning (EDL) ~\cite{sensoy2018evidential} is a probabilistic framework to model prediction uncertainty by employing evidential priors and subjective logic ~\cite{josang2016subjective}. In the EDL framework, the evidences $\boldsymbol{e}^v = [e^v_1,\cdots,e_K^v]$ for the $v$-th are obtained using Deep Neural Networks (DNNs), where the Softmax layer is replaced by a ReLU activation function.

Subjective logic provides a technical approach to associate the evidence $\boldsymbol{e}^v = [e^v_1,\cdots,e_K^v]$ with the parameters of the Dirichlet distribution $\boldsymbol{\alpha}^v = [\alpha_1^v,\cdots,\alpha_K^v]$. Specifically, each parameter $\alpha^v_k$ of the Dirichlet distribution is derived from $e^v_k$ as $\alpha^v_k = e^v_k + 1$. The belief mass $b^v_k$ and the uncertainty $u^v$ are then calculated as follows:
\begin{equation} 
b_k^v = \frac{e_k^v}{S^v} = \frac{\alpha_k^v - 1}{S^v} \quad \text{and} \quad u^v = \frac{K}{S^v},
\label{eq01}
\end{equation}
where $S^v = \sum\nolimits_{k=1}^{K} (e_k^v + 1) = \sum\nolimits_{k=1}^{K} \alpha^v_k$ denotes the Dirichlet strength. The projection probability is calculated by
\begin{equation}
    P_k^v = b_k^v + a_k^v u^v.
    \label{eq01_r1}
\end{equation}

From Eq. (\ref{eq01}), we can obtain an ordered triplet $\omega=(\boldsymbol{b},u,\boldsymbol{a})$ that reflects the multinomial opinion of an instance $\boldsymbol{x}^v_n$. Since the Dirichlet distribution parameter $\boldsymbol{a}$ is derived from the belief mass $\boldsymbol{b}$, a simplified representation of the multinomial opinion can be expressed as $\mathcal{M}=\{\boldsymbol{b},u\}$. Once multinomial opinions from multiple views are derived, designing an efficient opinion aggregation strategy becomes crucial for achieving trusted decision-making.

\subsection{Group Consensus Opinion Aggregation}

Social Identity Theory posits that individuals tend to value shared group opinions over personal viewpoints, which helps avoid conflict and fosters stable social environments \cite{hogg2016social}. Inspired by this theory, we argue that in a trusted decision-making process, partial aggregated group opinions should be prioritized over individual opinions. To this end, we propose a novel opinion aggregation strategy, termed Group Consensus Opinion Aggregation, which is detailed below.
 
%

\textbf{Definition 1.} \textbf{\textit{Group Consensus Opinion Aggregation.}} Without loss of generality, each view is assumed to be equally important. For $V$ opinions $\{\mathcal{M}^1,\mathcal{M}^2,\cdots,\mathcal{M}^V\}$, we aggregate them via

\begin{equation}
    \mathcal{M}^{\Diamond(V)} = \underbrace{\underbrace{\underbrace{\mathcal{M}^1 \oplus \mathcal{M}^2}_{\mathcal{M}^{\Diamond(2)}} \oplus \mathcal{M}^3}_{\mathcal{M}^{\Diamond(3)}} \oplus \dots \oplus \mathcal{M}^{V-1}}_{\mathcal{M}^{\Diamond(V-1)}} \oplus \mathcal{M}^V,
    \label{eq02}
\end{equation}
where $\mathcal{M}^{\Diamond(V)} = \{\boldsymbol{b}^{\Diamond(V)},u^{\Diamond(V)}\}$. Regarding the calculation of $\boldsymbol{b}^{\Diamond(V)}$ and $u^{\Diamond(V)}$, we present the aggregation process corresponding to their $k$-th components:
\begin{equation}
	\begin{aligned}
    b_k^{\Diamond(V)} &= \frac{e_k^{\Diamond(V)}}{S^{\Diamond(V)}}=\frac{\gamma^{\Diamond(V-1)} e_k^{\Diamond(V-1)} +\gamma^V e_k^V}{\gamma^{\Diamond(V-1)} S^{\Diamond(V-1)} + \gamma^V S^V},\\
    \end{aligned}
    \label{eq02_1}
\end{equation}
and 
\begin{equation}
	\begin{aligned}
	u^{\Diamond(V)}&=\frac{K}{\sum_{k=1}^{K}(e_k^{\Diamond(V)}+1)}\\
	&=\frac{K}{\sum_{k=1}^{K}(\gamma^{\Diamond(V-1)} e_k^{\Diamond(V-1)}+ \gamma^V e_k^V+1)},\\
	\end{aligned}
	\label{eq02_2}
\end{equation}
with $\gamma^{\Diamond(V-1)}$ and $\gamma^V$ being
\begin{equation}
    \gamma^{\Diamond(V-1)}=\frac{V-1}{V},\;\; \gamma^V=\frac{1}{V}.
    \label{eq05}
\end{equation}

From Eqs. (\ref{eq02_1}) to (\ref{eq05}), we see that the aggregated evidence $e_k^{\Diamond(V-1)}$ for the first $V-1$ views is assigned with a large weight $\gamma^{\Diamond(V-1)}$, whereas the evidence $e^V_k$ to be aggregated is assigned with a small weight $\gamma^V$. This indicates that, in the opinion aggregation process, the aggregated group opinion will play a dominant role. The further calculation results for Eq. (\ref{eq02_1}) and (\ref{eq02_2}) are given by
\begin{equation}
    b_k^{\Diamond(V)}=\frac{\gamma^{\Diamond(V-1)} b_k^{\Diamond(V-1)}u^V + \gamma^V b_k^V u^{\Diamond(V-1)}}{\gamma^V u^{\Diamond(V-1)}+ \gamma^{\Diamond(V-1)}u^V}
    \label{eq03}
\end{equation}
and 
\begin{equation}
    u^{\Diamond(V)} = \frac{u^{\Diamond(V-1)} u^V}{\gamma^V u^{\Diamond(V-1)}+\gamma^{\Diamond(V-1)}u^V}
    \label{eq04}
\end{equation}
The detailed derivation for the above equations and its effectiveness can be found in the supplementary material.


Our proposed aggregation strategy ensures that the group joint opinion is given higher priority than the individual subjective opinion. When dealing with conflicting opinions, our scheme tends to favor the group consensus opinion, thereby enhancing the realibity of decision-making.
%

\subsection{Uncertainty-based Sample Generation}

A common strategy for addressing the challenge of class imbalance is to generate new samples for the minority class to construct a balanced dataset. Among various techniques, the SMOTE method~\cite{chawla2002smote} is a well-established and straightforward approach that yields new samples by linearly interpolating between minority class samples and their nearest neighbors. While traditional SMOTE is effective in single-view scenarios, it is not directly applicable to multi-view settings. The primary challenge lies in the fact that SMOTE relies on neighbor analysis based on Euclidean distance calculations, which may yield inconsistent results across different views due to the inherent diversity among them. Specifically, the neighbor relationships identified in each view can differ significantly, potentially leading to conflicts in the synthetic samples generated from different views.


To overcome the aforementioned drawback, we attempt to introduce a distance metric to adapt SMOTE for multi-view scenarios. We build upon a natural premise: samples belonging to the same class exhibit consistency in their subjective opinions. Consequently, it is reasonable to infer that samples from the same category should also show similar evidence distributions. Based on this assumption, we develop a novel distance metric based on joint evidences. 


\textbf{Definition 2.} \textbf{\textit{Distance measures based on joint subjective evidence.}} Given any two samples $\{\boldsymbol{x}_i^v\}_{v=1}^V$ and $\{\boldsymbol{x}_j^v\}_{v=1}^V$, along with their corresponding joint opinions $\boldsymbol{e}_i$, $\boldsymbol{e}_j$, the distance between these two samples is defined as:
\begin{equation}
    D(\boldsymbol{x}_i,\boldsymbol{x}_j)=\sqrt{(e_{i1}-e_{j1})^2+\cdots+(e_{iK}-e_{jK})^2}.
    \label{eq06}
\end{equation}
In Eq. (\ref{eq06}), a smaller distance indicates a higher similarity between samples.

Apart from the limitation that the traditional SMOTE method cannot be applied to multi-view settings, another problem is that during the pseudo-data generation process, the weights for combining the center sample with its neighors are randomly selected, which may result in undesirable pseudo samples. To address this, we design an uncertainty-based sample generation scheme that ensures neighbors with low-uncertainty opinion are assigned with higher weights, thereby facilitating the generation of high-quality pseudo-samples. In the following, we present the details of our designed scheme.

For a minority class, we randomly select a data sample $\{\boldsymbol{x}_{\text{c}}^v\}_{v=1}^V$ as the center. The $R$ samples with the shortest distances to the center sample, calculated using Eq. (\ref{eq06}), are identified as its neighbors and denoted as $\{\boldsymbol{x}_1^v, \boldsymbol{x}_2^v, \dots, \boldsymbol{x}_R^v\}_{v=1}^V$. For the $v$-th view, the new pseudo sample is generated by
\begin{equation}
    \boldsymbol{x}_{new}^v =w_0^v\boldsymbol{x}_\text{c}^v+w_1^v\boldsymbol{x}_1^v+\cdots+w_R^v\boldsymbol{x}_R^v,
    \label{eq07}
\end{equation}
where $\sum_{r=0}^R w_r^v=1$ and $w_r^v\ge 0$. We expect that the weight $w^v_r$ to be assigned a high value if the aggregated opinion between the corresponding neighbor $\boldsymbol{x}^v_r$ and the center sample $\boldsymbol{x}^v_c$ exhibits low uncertainty, indicating that the relationship between this neighbor and the center sample is highly reliable. To achieve this objective, we begin with integrating the evidence of $\boldsymbol{x}^v_c$ with that of $\boldsymbol{x}^v_r$
\begin{equation}
    \boldsymbol{e}_{cr}^v = \frac{1}{2}\boldsymbol{e}_\text{c}^v+\frac{1}{2}\boldsymbol{e}_r^v,\;\; r=0,1,\cdots,R
    \label{eq08}
\end{equation}   
With the integrated evidence in Eq. (\ref{eq08}) and recalling (\ref{eq01}), we can obtain the corresponding opinions and projection probabilities for the neighbors, denoted as $\{\boldsymbol{b}_{cr}^v,u_{cr}^v\}_{r=0}^R$ and $\{\boldsymbol{P}^v_{cr}\}^R_{r=0}$, respectively. 

Here, we treat the projection probalibity as the prediction probability. This leads to a straightforward approach for aligning the distribution of the generated pseudo-samples with the true distribution, typically measured by cross-entropy. However, directly using cross entropy can not reflect the reliability of the generated samples. Thus, we introduce the influence of uncertainty and propose a novel criterion, termed Uncertainty Entropy. 

\textbf{Definition 3.} \textbf{\textit{Uncertainty Entropy.}} Given an opinion $\mathcal{M}=\{\boldsymbol{b}, u\}$ and the true label $\boldsymbol{y}$, the uncertainty entropy is defined as:
\begin{equation}
    H = -\text{exp}(u)\sum_{k=1}^K y_k \text{log}(b_k+ua_k),
    \label{eq09}
\end{equation}
where $\text{exp}(u)$ is the penalty coefficient. If the uncertainty $u=0$, then $\text{exp}(u)=1$, and the uncertainty has no effect on the entropy. Conversely, as the uncertainty $u$ increases, the entropy value also increases, indicating a greater discrepancy between the predicted classification and the true class.

With uncertainty entropy, we further present the following formulation for calculating $w^v_r$
\begin{equation}
    w_r^v = \frac{\mathcal{F}(H_{cr}^v)}{\mathcal{F}(H_c^v)+\sum_{r=1}^R\mathcal{F}(H_{cr}^v)}, \; r=0,1,\cdots,R  
    \label{eq10}
\end{equation}
where $\mathcal{F}(\cdot)$ is a monotonically decreasing function. If the uncertainty is low, from the above discussion, we know that the uncertainty entropy is small. Due to the monotonically decreasing property of $\mathcal{F}(\cdot)$, $\mathcal{F}(H_{cr}^v)$ will be assigned a large value. This implies that our scheme is capable of ensuring that neighbors with low-uncertainty opinions are assigned higher weights. In this paper, we use $\mathcal{F}(x)=x^{-1}$.

\subsection{Loss Function}


In this subsection, we introduce the loss function for taining our model. We first employ an adjusted cross-entropy loss function to ensure that all views generate appropriate non-negative view-specific evidence for classification:
\begin{equation} 
    \begin{aligned}  
        L_{ace}(\boldsymbol{\alpha}_n) &= \int \left[ \sum_{k=1}^{K}-y_{nk}\log p_{nk} \right] \frac{\Pi_{k=1}^K p_{nk}^{\alpha_{nk}-1}}{B(\alpha_n)}d_{P_n}\\   
        &=\sum_{k=1}^{K} y_{nk}(\psi(S_n) - \psi(\alpha_{nk}),
    \end{aligned}  
\end{equation}
where $\psi(\cdot)$ is the digamma function. 

Additionally, to ensure that incorrect classes in each sample produce lower evidence, we incorporate an extra KL divergence term into the loss function

\begin{equation}
    \begin{aligned} 
        L_{KL}(\boldsymbol{\alpha}_n) = & KL \left[D(p_n|\tilde{\alpha}_n)||D(p_n|1)\right]\\
        &= \log(\frac{\Gamma(\sum_{k=1}^{K} \tilde{\alpha}_{nk})}{\Gamma(K)\Pi^K_{k=1}\Gamma(\tilde{\alpha}_{nk})})\\
        & + \sum_{k=1}^{K}(\tilde{\alpha}_{nk}-1) \left[\psi(\tilde{\alpha}_{nk}) - \psi(\sum_{j=1}^{K}\tilde{\alpha}_{nj})\right],
    \end{aligned}  
\end{equation}
where $D(p_n|1)$ is the uniform Dirichlet distribution, $\tilde{\boldsymbol{\alpha}}_{n} = \boldsymbol{y}_n + (1-\boldsymbol{y}_n)\odot \boldsymbol{\alpha}_n$ is the Dirichlet parameters that reduces the incorrect classes to zero for the $n$-th instance, and $\Gamma(\cdot)$ is the gamma function. Therefore, given the Dirichlet distribution with parameter $\boldsymbol{\alpha}_n$ for the $n$-th instance, the loss is:
\begin{equation}
    L_{acc}(\boldsymbol{\alpha}_n) = L_{ace}(\boldsymbol{\alpha}_n) + \lambda_t L_KL(\boldsymbol{\alpha}_n),
\end{equation}
where $\lambda_t\in[0,1]$ is a balance factor with $t$ being the current training epoch index. The implementation procedures are summarized in Algorithm \ref{algorithm}.

\begin{algorithm}
    \caption{Implementation pseudocode of TMLC}
    \begin{algorithmic}
        \item[]\hspace*{-\algorithmicindent}\textbf{Input:} Multi-view dataset: $\{\boldsymbol{X}^v, \boldsymbol{y}\}_{v=1}^V$
         \item[]\hspace*{-\algorithmicindent}/---Train---/
         \item[]\hspace*{-\algorithmicindent}\textbf{Output:} Network's parameters.
        \STATE 01: \textbf{while} not converged \textbf{do}
        \STATE 02: \quad \textbf{for} $v=1:V$  \textbf{do}
        \STATE 03: \qquad $\textbf{e}^v$ $\leftarrow$ evidential network output.
        \STATE 04: \qquad Calculate opinion $\mathcal{M}^v$ by Eq. (1).
        \STATE 05: \quad \textbf{end for}
        \STATE 06: \quad Calculate joint opinion $\mathcal{M}^{\Diamond(V)}$ by Eq. (3).
        \STATE 07: \quad Calculate joint evidence $\boldsymbol{e}$ by Eq. (1).
        \STATE 08: \quad Calculate the joint distribution $\boldsymbol{\alpha}=\boldsymbol{e}+1$. 
        \STATE 09: \quad Calculate the overall loss $\mathcal{L}$ by Eq. (16).
        \STATE 10: \quad Update the networks by gradient descent by $\mathcal{L}$.
        \STATE 11: \textbf{end while}
        \STATE 12: Initialize pseudo-data set $\{\}$.
        \STATE 13: \textbf{while} minority class $k$-th not balanced \textbf{do}
        \STATE 14: \quad  Randomly select a center sample $\{\boldsymbol{x}_{\text{c}}^v\}_{v=1}^V$.
        \STATE 15: \quad Calculate multi-view neighbors by Eq. (9).
        \STATE 16: \quad \textbf{for} $v=1:V$  \textbf{do}
        \STATE 17: \qquad Generate pseudo date $\boldsymbol{x}_{new}^v$ by Eq. (10).
        \STATE 18: \quad \textbf{end while}
        \STATE 18: \quad Add $\left\{ \{\boldsymbol{x}_{new}^v\}_{v=1}^V,y_k=1\right\}$ to pseudo-data set.
        \STATE 19: \textbf{end for}
        \STATE 20: Add pseudo-data set to the training set.
        \STATE 21: Repeat training steps 01-11.
        \STATE 22: \textbf{return} networks parameters.
        \item[]\hspace*{-\algorithmicindent}/---Test---/
        \item[]\hspace*{-\algorithmicindent}\textbf{Output:}The final classification prediction and uncertainty. 
        \STATE 01: \textbf{for} $v=1:V$  \textbf{do}
        \STATE 02: \quad $\textbf{e}^v$ $\leftarrow$ evidential network output.
        \STATE 03: \quad Calculate opinion $\mathcal{M}^v$ by Eq. (1).
        \STATE 04: \textbf{end for}
        \STATE 05: Calculate joint opinion $\mathcal{M}^{\Diamond(V)}$ by Eq. (3).
        \STATE 06: \textbf{return} the decision $\boldsymbol{b}^{\Diamond(V)}$ and uncertainty $u^{\Diamond(V)}$.
    \end{algorithmic}
    \label{algorithm}
\end{algorithm}

\section{Experiment}


\subsection{Experimental Setup}
\textbf{Datasets.} \textbf{HandWritten}\footnotemark\footnotetext{https://archive.ics.uci.edu/ml/datasets/Multiple+Features} comprises 2000 instances of handwritten numerals ranging form ‘0’ to ‘9’, with 200 patterns per class. It is represented using six feature sets. \textbf{PIE}\footnotemark\footnotetext{http://www.cs.cmu.edu/afs/cs/project/PIE/MultiPie/MultiPie Home.html}  contains 680 instances belonging to 68 classes. We extract intensity, LBP, and Gabor as 3 views. \textbf{Caltech101}\footnotemark\footnotetext{http://www.vision.caltech.edu/Image Datasets/Caltech101} comprises 8677 images from 101 classes. We select 2386 samples from 20 classes, where each image has six features: Gabor, Wavelet Moments, CENTRIST, HOG, GIST, and LBP. \textbf{NUS-WIDE}\footnotemark\footnotetext{https://paperswithcode.com/dataset/nus-wide} consists of 269,648 images with 81 concepts. For the top 12 classes, we select 200 images from each class, with a total of 6 different views. \textbf{Scene15}\footnotemark\footnotetext{https://doi.org/10.6084/m9.figshare.7007177.v1} includes 4485 images from 15 indoor and outdoor scene categories. We extract three types of features GIST, PHOG, and LBP. \textbf{Animal}\footnotemark\footnotetext{http://attributes.kyb.tuebingen.mpg.de/} contains 37,322 images of 50 animal species with six features for each image. We select 11,673 samples contained 20 animal classes and four features for 4 views. 



\begin{table*}[h]
\centering
\small
\begin{tabular}{cccccccc}
\hline
\multirow{2}{*}{Methods} & \multicolumn{6}{c}{Datasets} \\ 
& HandWritten & PIE & Caltech101 & NUS-WIDE & Scene15 & Animal \\ \hline
TLC (2022) & 84.20$\pm$1.05 & 36.32$\pm$0.67 & 79.04$\pm$1.22 & 19.00$\pm$0.61 & 42.54$\pm$1.30 & 16.54$\pm$0.10 \\
H2T (2024) & 90.13$\pm$0.12 & 31.54$\pm$0.61 & 83.12$\pm$0.22 & 17.97$\pm$0.33 & 38.59$\pm$0.25 & 15.84$\pm$0.10 \\
TMC (2022) & 92.65$\pm$0.28 & 46.18$\pm$0.92 & 84.91$\pm$0.21 & 22.44$\pm$0.39 & 32.22$\pm$0.50 & 24.23$\pm$0.12 \\
ECML (2024) & 91.48$\pm$0.34 & \underline{47.28$\pm$0.74} & 82.87$\pm$0.56 & 22.35$\pm$0.27 & 41.31$\pm$0.71 & \underline{24.31$\pm$0.08} \\
DUA-Nets (2021) & 91.53$\pm$1.86 & 8.96$\pm$1.08 & 77.60$\pm$2.00 & 20.45$\pm$1.72 & 36.72$\pm$1.10 & 14.89$\pm$0.65 \\
DCP (2022) & \underline{93.48$\pm$1.33} & 4.56$\pm$0.64 & \underline{86.54$\pm$1.15} & \underline{29.04$\pm$2.21} & \underline{53.56$\pm$1.22} & 18.77$\pm$0.63 \\ 
UIMC (2023) & 91.35$\pm$0.46 & 45.88$\pm$1.51 & 79.58$\pm$0.74 & 20.98$\pm$0.80 & 48.02$\pm$0.38 & 20.90$\pm$0.41 \\ 
DMVLS (2024) & 81.28$\pm$1.05 & 11.62$\pm$1.18 & 72.03$\pm$1.29 & 21.48$\pm$1.19 & 30.19$\pm$0.88 & 18.63$\pm$0.81 \\ 
\cline{1-7}
Ours-v1  & 93.40$\pm$0.28 & 48.38$\pm$1.03 & 83.76$\pm$0.48 & 26.31$\pm$0.19 & 45.20$\pm$0.41 & 24.34$\pm$0.06 \\
Ours-v2 & 96.15$\pm$0.12 & 49.26$\pm$0.66 & 88.85$\pm$0.81 & 38.22$\pm$0.44 & 57.69$\pm$0.44 & 30.83$\pm$0.15 \\ 
Ours   & \textbf{96.25$\pm$0.32} & \textbf{62.43$\pm$0.70} & \textbf{89.48$\pm$0.24} & \textbf{38.96$\pm$0.37} & \textbf{58.57$\pm$0.37} & \textbf{31.35$\pm$0.20} \\ 
\hline
\end{tabular}
\caption{Accuracy (\%) on class-balanced test sets. The best results are highlighted in bold, while the second-best among baselines are underlined.}
\label{table:classficiation}
\end{table*}

\begin{table}[h]
\centering
\small
\begin{tabular}{c|c|cc}
\hline
\multirow{2}{*}{Datasets} & \multirow{2}{*}{Test sets} & \multicolumn{2}{c}{Methods} \\ \cline{3-4}
&  & ECML & Ours  \\ \hline
\multirow{2}{*}{HandWritten} & Normal & 97.00 & \textbf{97.50}  \\ \cline{2-4}
 & Conflictive & 92.69  & \textbf{94.80}  \\ \hline
\multirow{2}{*}{PIE} & Normal & 94.85 & \textbf{95.59}  \\ \cline{2-4}
 & Conflictive & 81.62  & \textbf{83.67}  \\ \hline
\multirow{2}{*}{Caltech101} & Normal & \textbf{93.88} & 93.08  \\ \cline{2-4}
 & Conflictive & 90.53  & \textbf{90.99}  \\ \hline
\multirow{2}{*}{NUS-WIDE} & Normal & 40.54 & \textbf{46.04}  \\ \cline{2-4}
 & Conflictive & 37.04  & \textbf{42.17}  \\ \hline
\multirow{2}{*}{Scene15} & Normal & 70.12 & \textbf{71.57}  \\ \cline{2-4}
 & Conflictive & 61.50  & \textbf{62.74}  \\ \hline
\multirow{2}{*}{Animal} & Normal & 58.31 & \textbf{59.48}  \\ \cline{2-4}
 & Conflictive & 49.39  & \textbf{51.05}  \\ \hline
\end{tabular}
\caption{Accuracy (\%) on normal and conflictive test sets. The best results are highlighted in bold.}
\label{table:conflict}
\end{table}

\textbf{Compared Methods.} The compared methods involve long-tail methods, trusted multi-view learning and other multi-view baselines. Specifically, 1) long-tail methods include: \textbf{TLC} (Trustworthy Long-tailed Classification)~\cite{li2022trustworthy} combines classification and uncertainty estimation in a multi-expert framework with dynamic expert engagement. \textbf{H2T} (Head-To-Tail)~\cite{li2024feature} augments tail classes by incorporating the diverse semantic information from head classes. 2) Trusted multi-view decision baselines include: \textbf{TMC} (Trusted Multi-view Classification)~\cite{han2022trusted} integrates different views at an evidence level using variational Dirichlet distribution and Dempster-Shafer theory. \textbf{ECML} (Evidential Conflictive Multi-view Learning)~\cite{xu2024reliable} puts forward a conflictive opinion aggregation model. 3) Multi-view classification baselines include: \textbf{DUA-Nets} (Dynamic Uncertainty-Aware Networks)~\cite{geng2021uncertainty} is an uncertainty aware method that uses reversal networks to integrate intrinsic information from different views into a unified representation. \textbf{DCP} (Dual Contrastive Prediction)~\cite{lin2022dual} enables consistency learning and missing view recovery by using an information-theoretical framework. \textbf{UIMC} (Uncertainty for Incomplete Multi-View Classification)~\cite{xie2023exploring} models the uncertainty of missing views through distribution construction and sampling and employs an evidence-based fusion strategy for trustworthy integration of imputed views. \textbf{DMVLS} (Dynamic Multi-view Labeling Strategy)~\cite{wan2024decouple} leverages shared and specific information with two classifiers to prioritize labeling low-confidence samples.

\textbf{Implementation Details.} Please see the supplementary material.

\begin{figure}[t]
  \centering
   \includegraphics[width=0.6\linewidth]{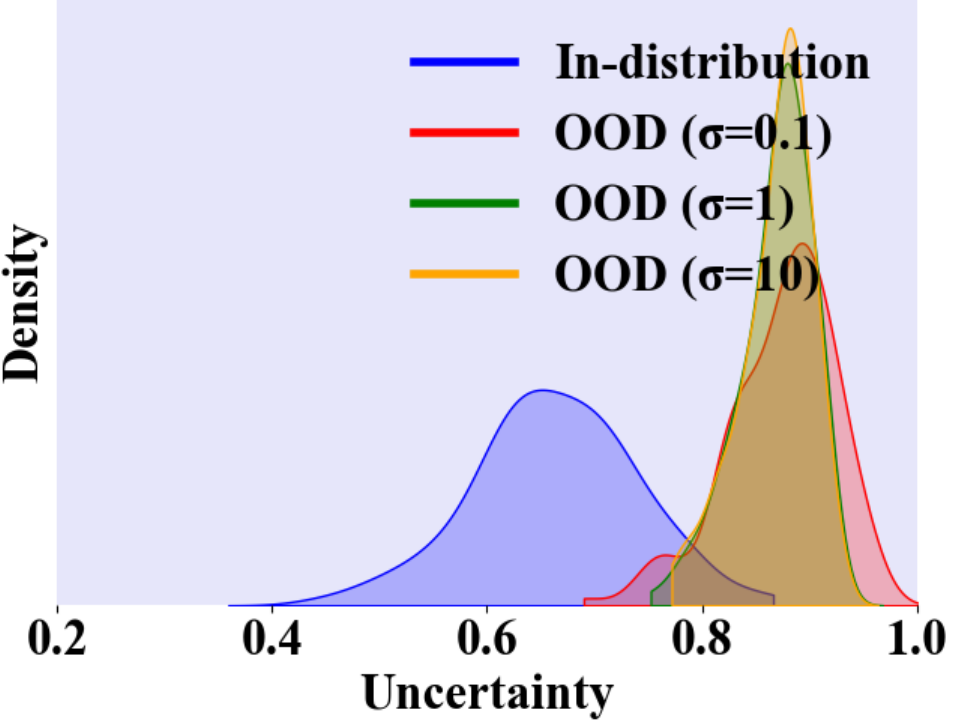}
   \caption{Density of overall uncertainty on the PIE
   dataset.}
   \label{fig:Uncertainty}
\end{figure}

\begin{figure}[t]
  \centering
   \includegraphics[width=0.6\linewidth]{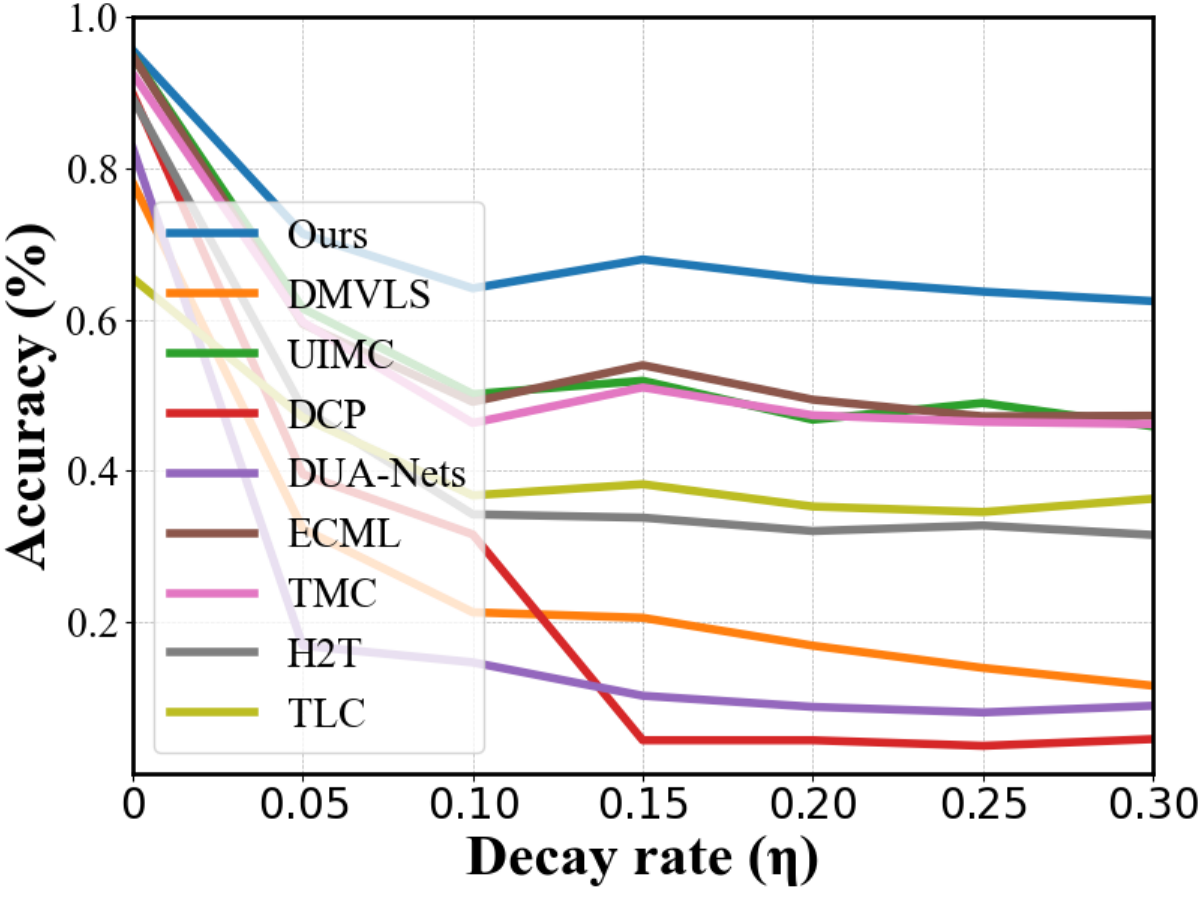}
   \caption{Performance comparison on PIE datasets with varying decay rates.}
   \label{fig:Different Decay Rates}
\end{figure}

\begin{figure}[t]
  \centering
   \includegraphics[width=0.6\linewidth]{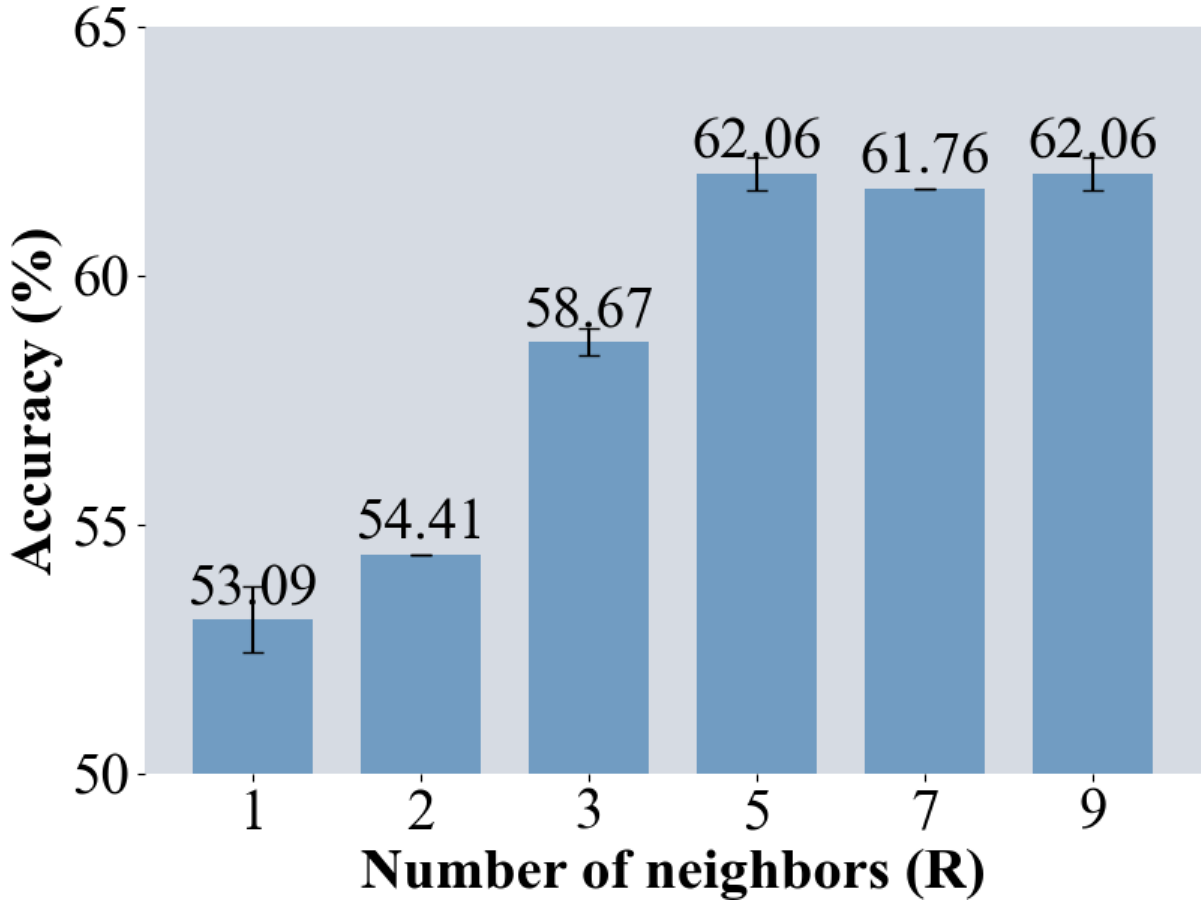}
   \caption{Accuracy comparison with respect to $R$.}
   \label{fig:R_PIE}
\end{figure}

\begin{figure*}[t]
  \centering
   \includegraphics[width=0.75\linewidth]{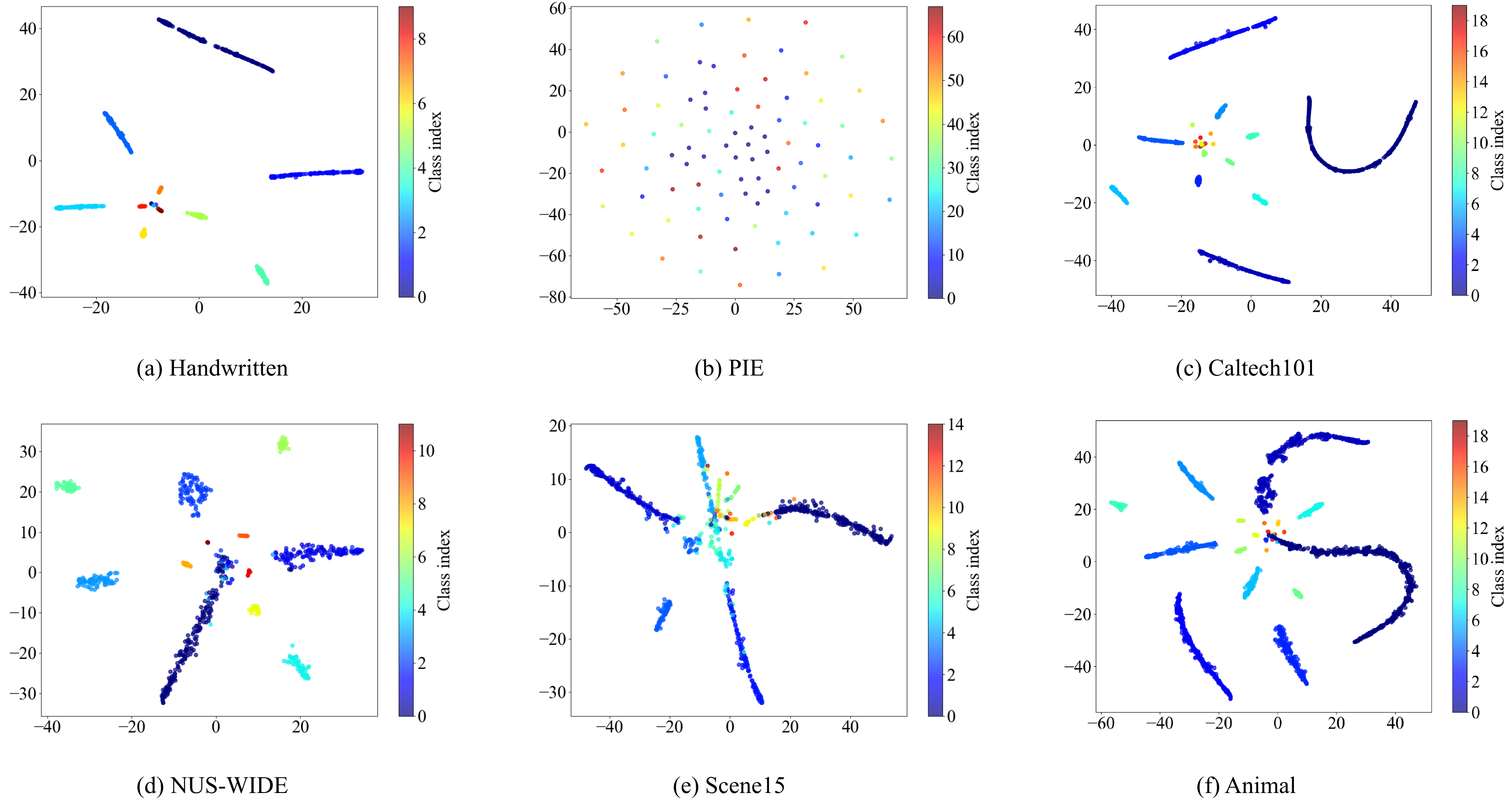}
   \caption{Joint evidence visualization on different datasets.}
   \label{fig:evidence visual}
\end{figure*}

\subsection{Experiment Results}

\textbf{Accuracy Comparison.} Table \ref{table:classficiation} presents the comparison results of various methods on standard test sets. It is evident that our proposed TMLC method significantly outperforms all baseline methods on the class-balanced test set, particularly on the PIE, NUS-WIDE, and Animal datasets, where it achieves performance improvements exceeding 5\%. We attribute this enhancement primarily to our uncertainty-based sample generation scheme, which ensures that the generated pseudo-samples more closely approximate the true sample distribution. These pseudo-samples effectively mitigate the class imbalance issue in multi-view data, thereby improving the model's classification accuracy. To validate the effectiveness of the generated pseudo-samples, we conduct a comparative experiment by excluding the pseudo-sample generation module, resulting in a variant labeled as Ours-v1. Furthermore, to verify the superiority of the weight calculation scheme, we replace the designed weight mechanism with random weights, resulting in a variant labeled as Ours-v2. The results show that Ours-v1 performs worse than the complete model across all tested datasets, demonstrating that the uncertainty-based sample generation scheme plays a crucial role in enhancing the network's classification performance. It is also seen that Ours-v2 exhibits inferior performance relative to the complete model, confirming that the uncertainty-induced weight scheme more effectively integrates neighboring samples from different views, thereby generating synthetic samples that better align with the characteristics of the minority class.

\textbf{Performance for Conflictive Noise.} To validate the superiority of the proposed group consensus opinion aggregation mechanism, we conduct experiments by introducing conflictive noise into the test data \cite{xu2024reliable}. We compare our approach with ECML, a state-of-the-art (SOTA) baseline for handling conflictive instances. Table \ref{table:conflict} summarizes the comparison results on six datasets under both normal and conflict noise conditions. The results demonstrate that, in both scenarios, the group consensus opinion aggregation mechanism generally outperforms ECML. Specifically, on the NUS-WIDE dataset, our method achieves performance improvements of 5.5\% and 5.13\% in normal and conflict noise settings, respectively. These findings highlight the effectiveness of our proposed group consensus opinion aggregation in enhancing the aggregation of multi-view opinions, thereby leading to more reliable classification results.

\textbf{Uncertainty Estimation.} In order to assess the uncertainty of the model predictions, we visualize the distribution of both in-distribution and out-of-distribution samples. Specifically, the original samples and the samples with added Gaussian noise are considered as in-distribution and out-of-distribution samples, respectively. The experiment is conducted on the PIE dataset, where Gaussian noise with different standard deviations $(\sigma = 0.1, \sigma = 1, \text{and } \sigma = 10)$ is added to the test samples, as shown in Fig. \ref{fig:Uncertainty}. The results indicate that as the noise intensity increases, the uncertainty of the data also increases, and the distribution curve of the out-of-distribution samples deviates more significantly from that of the in-distribution samples. This suggests that the proposed group consensus opinion aggregation strategy effectively captures the uncertainty, thereby improving the overall reliability of the model.

\textbf{Performance Evaluation under Different Decay Rates.} To demonstrate the superiority of our TMLC model in noisy environments, we conduct experiments by introducing varying levels of decay rates into the training data, with the decay rate ranging over $[0, 0.05, 0.10, 0.15, 0.20, 0.25, 0.30]$. Fig. \ref{fig:Different Decay Rates} illustrates the performance comparison of all methods on the PIE dataset under different decay rates. It is evident that, compared to the baseline methods, our TMLC model consistently outperforms them across all tested datasets at each decay rate, further validating the effectiveness of the proposed pseudo-data generation scheme.

\textbf{Parameter Selection.} Since the model requires a predefined number of neighbors, $R$, we conduct experiments with different values of $R = [1, 2, 3, 5, 7, 9]$ to investigate its impact. As shown in Fig. \ref{fig:R_PIE}, as the number of neighbors increases, the generated multi-view pseudo-samples incorporate more relevant sample information, which enhances the diversity of the samples and improves the classification performance. However, beyond a certain value of $R$, the classification performance approaches stability. This is because the sample knowledge has already been sufficiently integrated, and further increasing the number of neighbors does not provide additional useful information, resulting in negligible improvements in classification performance.

\textbf{Joint Evidence Visualization.} To validate the rationale of the newly defined evidence-based distance metric in the uncertainty-based sample generation scheme, Fig. \ref{fig:evidence visual} visualizes the evidence obtained from the aggregation mechanism on six multi-view datasets. Our observations are as follows: (1) Evidence points belonging to the same class are clustered closely together, while those from different classes are distinctly separated. This demonstrates that the evidence-based distance effectively captures class-specific information and accurately constructs neighbor relationships. (2) As the degree of class imbalance increases, the limited number of training samples results in reduced evidence and heightened uncertainty. Consequently, these classes tend to form neighbor relationships with other challenging-to-classify classes.

\section{Conclusion}

In this paper, we propose a trusted multi-view learning framework for long-tailed classification, namely TMLC . TMLC makes two significant contributions. First, inspired by Social Identity Theory, we design a group consensus opinion aggregation mechanism that aligns decision-making with the collective judgment of the majority. Second, we develop an innovative pseudo-data generation strategy that integrates a novel evidence-based distance metric and an uncertainty-guided module to produce high-quality synthetic samples. Extensive experiments demonstrate the superior performance of TMLC.

\section{Acknowledgments}

This work was supported by the National Natural Science Foundation of China under Grant No.62201475 and No.62406241, Sichuan Science and Technology Program under Grant No.2024NSFSC1436.

\appendix

\bibliography{aaai2026}

\clearpage
\twocolumn[
\begin{center}
    {\LARGE \bfseries Supplementary Materials}
\end{center}
\vspace{3em}
]

\maketitle


In this supplementary material, we provide a detailed derivation of the group consensus opinion aggregation mechanism, demonstrate the effectiveness of opinion consensus aggregation, detail the implementation details, and present supplementary experimental results.

\section{Derivation of Group Consensus Opinion Aggregation}

This part provides a detailed calculation for \emph{group consensus opinion aggregation}. Specifically, the details of deriving the aggregated opinion $\mathcal{M}^{\Diamond}=\{{\boldsymbol{b}^{\Diamond},u^{\Diamond}}\}$ will be shown below. Before proceeding, we need to recall some basic formulas in the EDL. The belief mass $b_k$ and uncertainty $u$ are calculated by:
\begin{equation}
    b_k=\frac{e_k}{S},u=\frac{K}{S},S=\sum_{k=1}^K(e_k+1)
    \label{eq:1}
\end{equation}
Assume that $\boldsymbol{e}^A$ and $\boldsymbol{e}^B$ are two evidences to be aggregated, and let $\gamma^A$ and $\gamma^B$ denote the corresponding weights of the two opinions in the fusion process. Then the aggregated evidence $\boldsymbol{e}^{\Diamond}$ is computed as:
\begin{equation}
        e_K^\Diamond = \gamma^A e_k^A + \gamma^B e_k^B \quad (\gamma^A+\gamma^B=1)
        \label{eq:2}
\end{equation}

With these preliminaries in place, we now dervie $\mathcal{M}^{\Diamond}=\{{\boldsymbol{b}^{\Diamond},u^{\Diamond}}\}$. According to the above Eqs. (\ref{eq:1}) and (\ref{eq:2}), the aggregated opinion $\mathcal{M}^{\Diamond}=\{{\boldsymbol{b}^{\Diamond},u^{\Diamond}}\}$ is calculated by:
\begin{equation}
\begin{aligned}
	u^{\Diamond} &=\frac{K}{S^{\Diamond}} =\frac{K}{\sum_{k=1}^{K}(e_k^{\Diamond}+1)}\\
	&= \frac{K}{\sum_{k=1}^{K}(\gamma^A e_k^A+ \gamma^B e_k^B+1)}
\end{aligned}
\label{eq:3}
\end{equation}

\noindent Due to $\gamma^A+\gamma^B=1$, from Eq. (\ref{eq:3}) we have
\begin{equation}
\begin{aligned}
	u^{\Diamond} &= \frac{K}{\sum_{k=1}^{K}(\gamma^A (e_k^A+1)+ \gamma^B (e_k^B+1))}\\
	&= \frac{K}{\gamma^A\sum_{k=1}^{K} (e_k^A+1)+ \gamma^B\sum_{k=1}^K (e_k^B+1)}
\end{aligned}
\label{eq:4}
\end{equation}

Substituting $S=\sum_{k=1}^K(e_k+1)$ into Eq. (\ref{eq:4}) and perform some algebraic operations, we finally arrive at
\begin{equation}
	u^{\Diamond} = \frac{u^A u^B}{\gamma^A u^B +\gamma^B u^A}
	\label{eq:5}
\end{equation}
This completes the derivation for $u^{\Diamond}$. 

Next, we focus on the derivation of $b_k^\Diamond$. With the definition of $b_k$ in Eq. (\ref{eq:1}) and the definition of $e_k$ in Eq. (\ref{eq:2}), we obtain
\begin{equation}
\begin{aligned}
	b_k^\Diamond &= \frac{e_k^\Diamond}{S^\Diamond} \\
    &=\frac{\gamma^A e_k^A +\gamma^B e_k^B}{\gamma^A S^A + \gamma^B S^B}
\end{aligned}
\label{eq:6}
\end{equation} 

\noindent By using a trick that $\frac{K S^A S^B}{K S^A S^B}=1$, the above equation is further written as
\begin{equation}
\begin{aligned}
	b_k^\Diamond & =\frac{\gamma^A e_k^A +\gamma^B e_k^B}{\gamma^A S^A + \gamma^B S^B}  \cdot \frac{K S^A S^B}{K S^A S^B} \\
    & =\frac{K(\gamma^A e_k^A +\gamma^B e_k^B)}{S^A S^B} \cdot \frac{S^A S^B}{K(\gamma^A S^A + \gamma^B S^B)} \\
    & =(\gamma^A \frac{e_k^A}{S^A} \cdot \frac{K}{S^B} + \gamma^B \frac{e_k^B}{S^B} \cdot \frac{K}{S^A}) \cdot \frac{S^A S^B}{K(\gamma^A S^A + \gamma^B S^B)} \\
\end{aligned}
\label{eq:7}
\end{equation}

\noindent With the definition of $u$ in Eq. (\ref{eq:1}), and after some algebraic operations, we finally have
\begin{equation}
\begin{aligned}
	b_k^\Diamond = \frac{\gamma^A b_k^A u^B + \gamma^B b_k^B u^A}{\gamma^A u^B + \gamma^B u^A}
\end{aligned}
\end{equation}
This completes the derivation for $b_k^\Diamond$.

\section{Effectiveness Illustration of Opinion Consensus Aggregation}
To illustrate that our Group Consensus Opinion Aggregation can effectively aggregate the uncertainty of two opinions, we make the following changes to the aggregated uncertainty $u^{\Diamond}$:

\begin{align}
    u^\Diamond  &= \frac{u^A u^B}{\gamma^A u^B +\gamma^B u^A} \notag\\
    &= \frac{u^A u^B}{\gamma^A u^B +\gamma^B u^A} \cdot 1 \notag\\
    &= \frac{u^A u^B}{\gamma^A u^B +\gamma^B u^A} \cdot \frac{u^A}{u^A} \notag\\
    &= \frac{u^A u^B}{\gamma^A u^A u^B +\gamma^B u^A u^A} \cdot u^A \notag\\
    &=\frac{1}{\gamma^A + \gamma^B \frac{u^A}{u^B}} \cdot u^A 
\end{align}

Let $u^B$ is the uncertainty of the new opinion, $u^A$ is the uncertainty of the original opinion, the uncertain mass of the aggregated opinion and the original opinion is:
\begin{equation}
    \begin{cases}
    \textbf{if} \; u^A>u^B,\textbf{then} \; \gamma^A+\gamma^B\frac{u^A}{u^B} > \gamma^A+\gamma^B =1,u^\Diamond<u^A \\
    \textbf{if} \; u^A=u^B,\textbf{then} \; \gamma^A+\gamma^B\frac{u^A}{u^B} = \gamma^A+\gamma^B =1,u^\Diamond=u^A \\
    \textbf{if} \; u^A<u^B,\textbf{then} \; \gamma^A+\gamma^B\frac{u^A}{u^B} < \gamma^A+\gamma^B =1,u^\Diamond>u^A
\end{cases}
\end{equation}
We can find that if the uncertain mass of the new opinion $u^B$ is smaller than the original uncertain mass $u^A$, the uncertain mass of the aggregated opinion $u^\Diamond$ would be smaller than the original one; conversely, it would be larger.

\begin{table*}[h!]
\centering
\caption{Performance comparison on multiple datasets (\%)}
\begin{tabular}{l l c c c | l l c c c}
\toprule
Dataset & Method & Head & Med & Tail & Dataset & Method & Head & Med & Tail \\
\midrule
        & TLC  & \underline{99.51} & 95.70 & 24.26 &       & TLC  & 33.18 & 22.93 & 7.82 \\
        & TMC  & 98.85 & 95.70 & 24.92 &       & TMC  & 40.32 & 21.01 & 4.62 \\
Caltech101 & DCP  & 98.31 & \underline{96.98} & \underline{49.18} & NUS-WIDE & DCP  & 38.57 & \underline{25.96} & \underline{26.15} \\
        & DMVLS & 99.24 & 79.34 & 9.84 &       & DMVLS & \textbf{47.93} & 23.53 & 4.10 \\
        & Ours & \textbf{100.00} & \textbf{99.14} & \textbf{58.40} &       & Ours & \underline{47.62} & \textbf{41.11} & \textbf{41.54} \\
\midrule
Dataset & Method & Head & Med & Tail & Dataset & Method & Head & Med & Tail \\
\midrule
        & TLC  & 89.95 & 42.46 & 11.26 &       & TLC  & 56.57 & 19.76 & 2.84 \\
        & TMC  & 83.98 & 7.51 & 0.48 &       & TMC  & 70.46 & \underline{25.67} & 1.66 \\
Scene15 & DCP  & 90.26 & \textbf{68.83} & \underline{22.04} & Animal & DCP  & 60.83 & 21.58 & \underline{4.12} \\
        & DMVLS & \underline{92.78} & 45.76 & 1.13 &       & DMVLS & \underline{75.08} & 20.58 & 0.03 \\
        & Ours & \textbf{98.12} & \underline{55.02} & \textbf{28.53} &       & Ours & \textbf{81.12} & \textbf{48.41} & \textbf{9.50} \\
\bottomrule
\end{tabular}
\label{table:comparison}
\end{table*}

\section{Datasets Characteristics and Implementation Details}

Table \ref{table:dataset} shows the details of datasets.

\begin{table}[htbp]
\centering
\begin{tabular}{|c|c|c|c|}
\hline
\textbf{Dataset} & \textbf{Size} & \textbf{$K$} & \textbf{Dimensionality} \\ \hline
HandWritten & 2000 & 10 & 240/76/216/47/64/6 \\ \hline
PIE & 680 & 68 & 484/256/279 \\ \hline
Caltech101 & 2386 & 20 & 48/40/254/1984/512/928 \\ \hline
NUS-WIDE & 2400 & 12 & 64/144/73/128/225/500 \\ \hline
Scene15 & 4485 & 15 & 20/59/40 \\ \hline
Animal & 11673 & 20 & 2689/2000/2001/2000 \\ \hline
\end{tabular}
\caption{Dataset summary.}
\label{table:dataset}
\end{table}

\begin{figure}[t]
  \centering
   \includegraphics[width=0.7\linewidth]{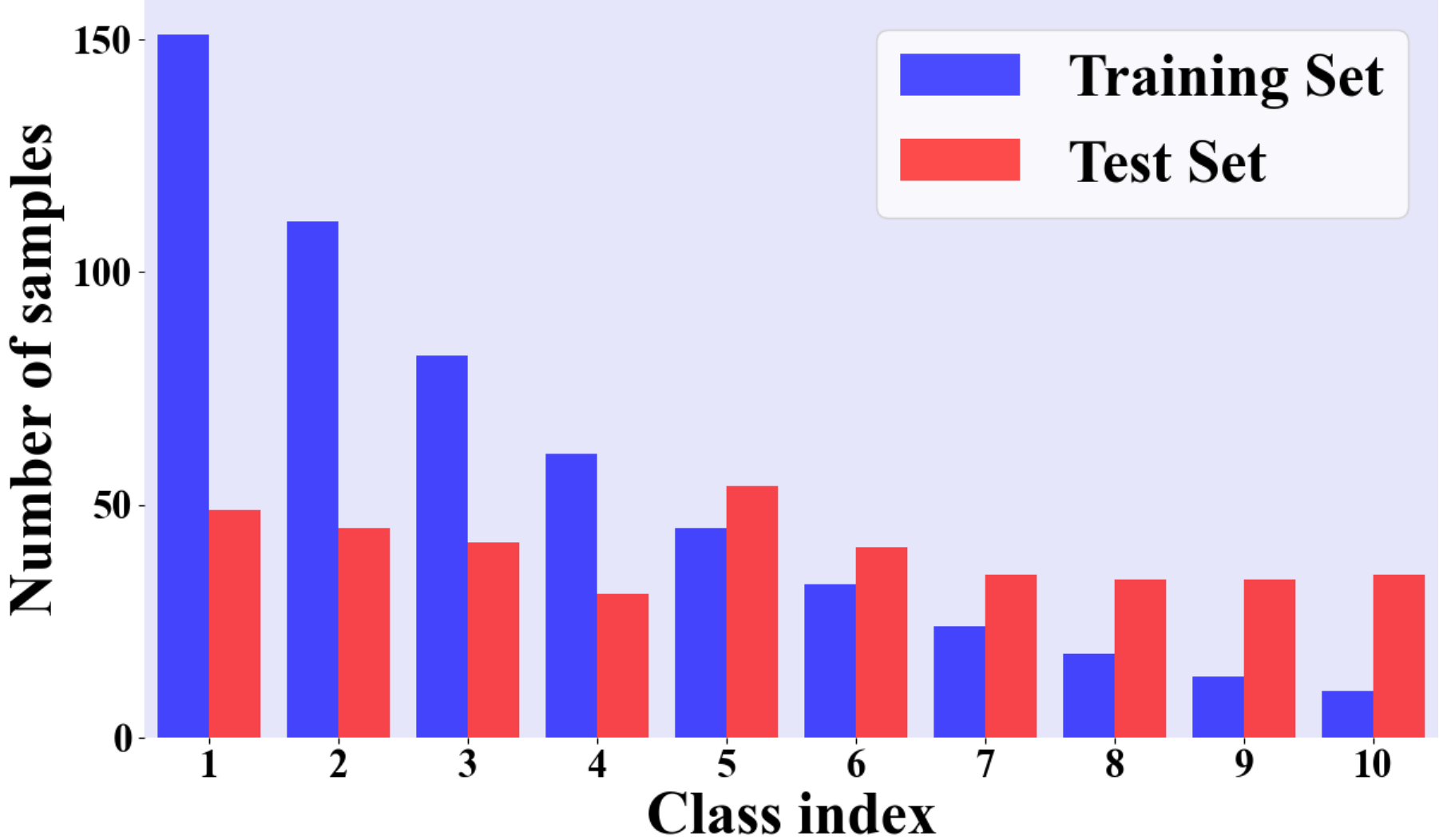}
   \caption{Sample distribution of training and test sets on HandWritten dataset.}
   \label{fig:Number of samples}
\end{figure}

\textbf{Implementation Details.} To generate a class-imbalanced training set and a class-balanced testing set, we conduct the following operations: (1) We randomly select 80\% of the samples for training and use the remaining 20\% for testing, thereby creating the standard training and test sets. (2) To further construct a long-tailed training set from the standard training set, we employ a widely adopted strategy~\cite{li2022trustworthy}. Specifically, we sample a subset using a specific decay distribution $P_{d}(y)$, such as an exponential distribution~\cite{cui2019class} or a Pareto distribution~\cite{liu2019large}, as follows:
\begin{equation}
    P^{'}(\{\boldsymbol{X}^{v}\}^V_{v=1} | y=k) 
    = P_{d}\{y=k\}P(\{\boldsymbol{X}^{v}\}^V_{v=1} | y=k).
\end{equation}
where samples from the $k$-th class are randomly selected with probability $P_{d}\{y=k\}$ to construct a class imbalance multi-view classification input. In our experiment, we employ an exponential decay rate $\eta=0.3$ to generate the imbalanced training set, where the number of samples decreases as the class label increases. Fig. \ref{fig:Number of samples} shows the sample distribution of traning and test sets using our data generation strategy on the HandWritten dataset. It is clear that the the training set follows a long-tail distribution, with tail classes containing significantly fewer samples than head classes. For single-view methods, we report the results from the best-performing view. All results are obtained by averaging 10 trials.

\section{Supplementary Experiments}

To further evaluate the effectiveness of our model under long-tailed distributions, we conduct a fine-grained accuracy analysis on head, medium, and tail classes across four benchmark datasets: Caltech101, NUS-WIDE, Scene15, and Animal. For each dataset, classes are sorted in descending order based on the number of training samples. The top three classes with the largest sample sizes are defined as head, the next five as medium, and the remaining classes as tail. This stratification allows for a more comprehensive assessment of model performance across different frequency regimes.

We compare our method against four representative baselines: TLC, TMC, DCP, and DMVLS. As shown in Table~\ref{table:comparison}, our model achieves superior performance on the majority of head, medium, and tail subsets across all datasets. Notably, the performance gap is most pronounced in the tail region, where our method consistently outperforms existing approaches by a significant margin—demonstrating its enhanced generalization capability for underrepresented classes.

As expected, all methods exhibit a performance drop on tail classes due to the extreme sample scarcity, which poses a fundamental challenge for reliable model learning. However, our approach mitigates this issue more effectively, maintaining higher accuracy even in the most data-deficient categories. This highlights its robustness and balanced learning dynamics across class frequencies.



\end{document}